%% file: root.tex
\documentclass[letterpaper, 10 pt, conference]{ieeeconf}  %

\IEEEoverridecommandlockouts                              %

\usepackage{graphics} %
\usepackage{epsfig} %
\usepackage{times} %
\usepackage{amsmath} %
\usepackage{amssymb}  %
\usepackage{nccmath}
\usepackage{mathtools}
\usepackage{nicefrac}
\usepackage{xcolor}
\usepackage{hyperref}
\usepackage{booktabs}
\usepackage[labelformat=simple]{subcaption}
\usepackage{siunitx}
\usepackage{arydshln}

\usepackage[font=small]{caption}
\def\equationautorefname~#1\null{Eq.~(#1)\null}

\newcommand{\appref}[1]{\hyperref[#1]{Appendix~\ref*{#1}}}

\DeclareMathOperator*{\EXP}{\mathbb{E}}%
\newcommand{\s}{s}
\newcommand{\pit}{\pi_{\theta}}
\newcommand{\pip}{\pi_{\text{prior}}}
\newcommand{\pib}{\pi_{b}}
\newcommand{\pie}{\pi_{\text{teacher}}}
\newcommand{\B}{\mathcal{B}}
\newcommand{\Dataset}{\mathcal{D}_{\text{dataset}}}
\newcommand{\D}{\mathcal{D}}

\newcommand{\KLcompact}[2]{D_{\text{KL}}\!\left(#1\middle\|#2\right)}

\title{\LARGE \bf
How to Spend Your Robot Time: \\
Bridging Kickstarting and Offline Reinforcement Learning  \\
for Vision-based Robotic Manipulation
}

\author{
    Alex X. Lee$^{*}$,
    Coline Devin$^{*}$,
    Jost Tobias Springenberg$^{*}$,\\
    Yuxiang Zhou,
    Thomas Lampe,
    Abbas Abdolmaleki,
    Konstantinos Bousmalis
    \\[1mm]
    DeepMind, London, UK
    \thanks{$^{*}$ Equal contribution.}
}

\begin{document}

\maketitle
\thispagestyle{plain}
\pagestyle{plain}
\begin{abstract}

Reinforcement learning (RL) has been shown to be effective at learning control from experience. However, RL typically requires a large amount of online interaction with the environment. This limits its applicability to real-world settings, such as in robotics, where such interaction is expensive. In this work we investigate ways to minimize online interactions in a target task, by reusing a suboptimal policy we might have access to, for example from training on related prior tasks, or in simulation. To this end, we develop two RL algorithms that can speed up training by using not only the action distributions of teacher policies, but also data collected by such policies on the task at hand. We conduct a thorough experimental study of how to use suboptimal teachers on a challenging robotic manipulation benchmark on vision-based stacking with diverse objects. We compare our methods to offline, online, offline-to-online, and kickstarting RL algorithms. By doing so, we find that training on data from both the teacher and student, enables the best performance for limited data budgets. We examine how to best allocate a limited data budget -- on the target task -- between the teacher and the student policy, and report experiments using varying budgets, two teachers with different degrees of suboptimality, and five stacking tasks that require a diverse set of behaviors. Our analysis, both in simulation and in the real world, shows that our approach is the best across data budgets, while standard offline RL from teacher rollouts is surprisingly effective when enough data is given.
\end{abstract}

\vspace{1mm}
\section{Introduction}
Learning-based methods have enabled the field of robotics to tackle a wider range of applications, particularly when operating directly from high-dimensional sensor inputs and in difficult-to-model environments.
Reinforcement Learning (RL) is an appealing approach as it can discover novel solutions for a given task.
However, online RL, i.e. RL that learns by interacting with the world while learning, is difficult to use on real robots due to the time required to continuously collect new experience, and the large-data requirements of modern learning algorithms. This is especially true for image-based, long-horizon tasks with only sparse rewards.

\begin{figure}[t]
    \centering
    \includegraphics[width=\columnwidth,trim={30px 30px 30px 20px},clip]{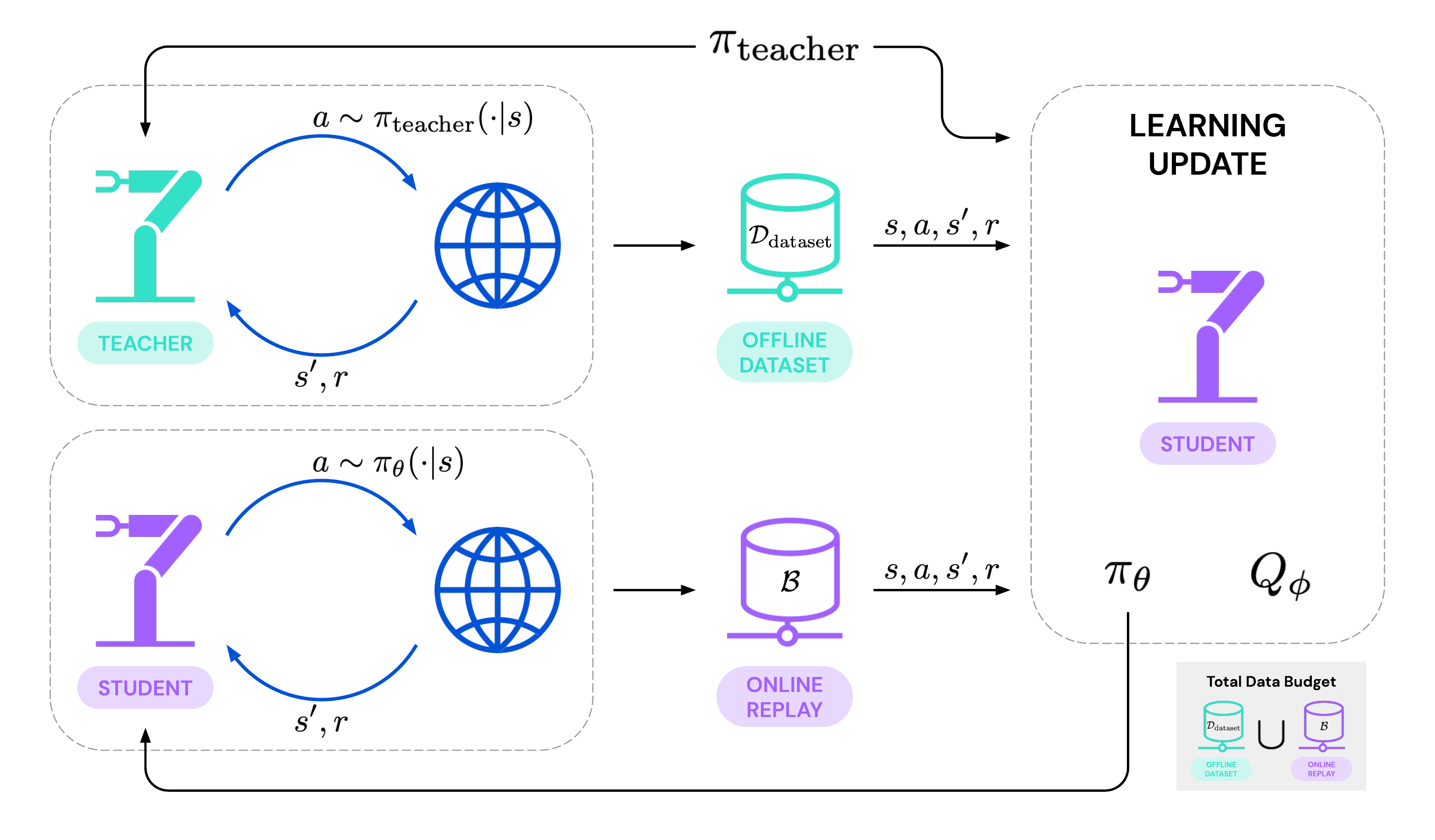}
    \caption{Our problem setting: the algorithm is given access to the environment and a suboptimal teacher  $\pie$.
    We control the data budget -- the total number of episodes collected online by the student $\pit$ and the ones collected offline by the teacher.
    Our methods are actor-critic algorithms that can choose how to split the data budget between teacher and student episodes, while having access to the teacher for action supervision in the learning update.
    }
    \label{fig:policy_finetuning_framework}
\end{figure}

In this work, we investigate how to best utilize the limited number of interactions possible on real robots, by leveraging existing policies that are suboptimal. Such policies are often available from training on prior related tasks or in simulation. 
This general setting where the learner is given access to the environment and, in addition, has access to a reference policy is called \textit{policy finetuning}. Not to be confused with the setting of finetuning the parameters of a teacher through further training, policy finetuning has recently been theoretically examined by Xie et al.~\cite{xie2021policyfinetuning}, who also coined the term.
The reference policy, which we refer to as the \textit{teacher}, can be queried at any given state and does not need to be parametric.
The policy finetuning setting  is flexible and encapsulates existing research areas widely studied in recent years. These are: (a) \textit{online RL} where the student learns from its own interactions and ignores the teacher, (b) \textit{offline RL} from a dataset collected by the teacher in the target task, (c)  \textit{offline-to-online RL}, where the goal is to quickly adapt a policy which has been pre-trained offline; and (d) \textit{kickstarting} or \textit{DAgger-style~\cite{ross2011dagger} learning}, where the goal is to speed up online RL by using feedback from the teacher. 

\begin{figure}%
    \centering
    \includegraphics[width=0.48\textwidth,trim={3mm 0 0 0},clip]{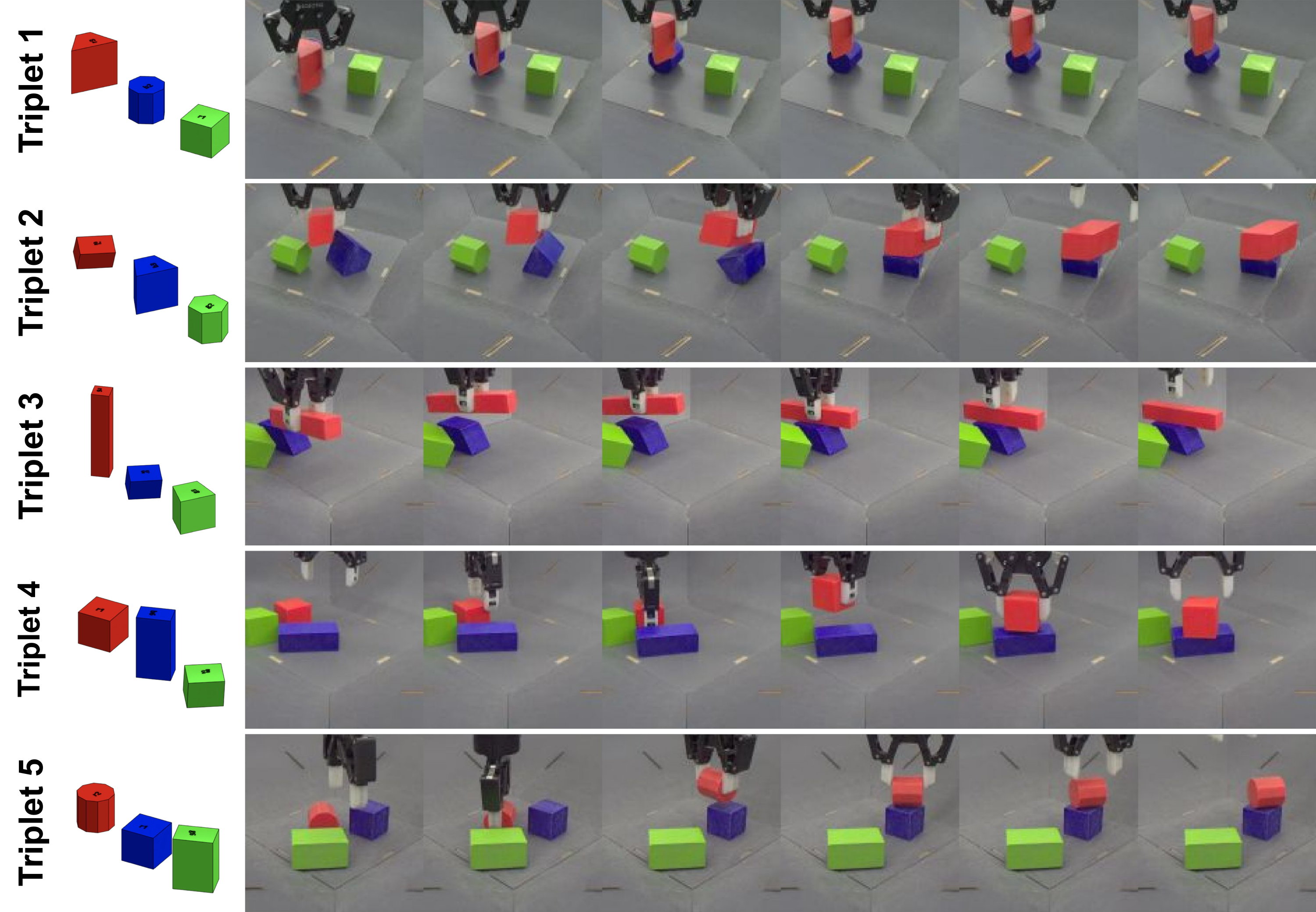} 
    \caption{The 5 RGB-stacking task variations in our experiments~\cite{lee2021beyond}. For each of these triplets, the task is to stack the red object on the blue one, while the green one is a distractor. %
    Each variation poses a unique challenge: Triplet~1 requires a precise grasp of the top object; Triplet~2 often requires the top object to be used as a tool to flip the bottom object before stacking; Triplet~3 requires balancing; Triplet~4 requires precision stacking (the object centroids need to align); and the top object of Triplet~5 can easily roll off.
    }
    \label{fig:triplet_challenges}
\end{figure}

We here focus on a policy finetuning setting in which we use both feedback from the teacher and data collected by it in the target task.
We empirically investigate different algorithmic choices within this problem setting (illustrated in \autoref{fig:policy_finetuning_framework}), including how much of the data for training should come from the teacher and how to use the teacher for action supervision.
We focus our experiments on a challenging robotic manipulation benchmark, RGB-Stacking~\cite{lee2021beyond}. 
This benchmark consists of a variety of image-based, sparse-reward stacking tasks for objects with diverse shapes that each require distinct behaviors.
We perform experiments on five such stacking tasks (shown in \autoref{fig:triplet_challenges}), and with two teacher policies with different degrees of suboptimality. %
Our contribution is twofold: i) we present a systematic approach to modifying online and offline RL algorithms for the general policy finetuning setting by training on interaction data from both the teacher and the online student,  while also leveraging the teacher for action supervision, and ii) we thoroughly investigate how our new methods perform against 7 baselines over different data budgets in simulation and on real robots.

\section{Related Work}
Our work in this paper investigates the merits of algorithms that fall on the spectrum of the policy finetuning problem setting, including offline RL and kickstarting. While a full survey of these is out of scope, we highlight connections between the algorithms we consider and prior work.

The literature on offline RL tackles the problem of learning a policy from a fixed dataset of transitions. The main problem in this setting is that standard off-policy RL algorithms -- e.g. naive policy optimization against a learned Q-function -- suffer from the problem of policies exploiting extrapolation errors in the learned action-value function approximators~\cite{fujimoto2019bcq,kumar2019bear}. 
Recent algorithms have attempted to fix such issues by either incorporating a pessimism bias for unseen actions into the Q-function e.g. \cite{kumar2020cql,kostrikov2021iql,ghasemipour2021emaq} or by constraining the policy optimization to stay close to the samples present in the data e.g. \cite{nair2020awac,wang2020crr,siegel2019keep,fujimoto2021minimalist,wu2019brac,xu2021sbac}. 
While most studies only consider learning over simulated environments and low-dimensional state features,
recent studies that do show successful learning from high-dimensional observations (such as images) are actor-critic algorithms with constraints or regularization on the policy~\cite{wang2020crr,nair2020awac,fujimoto2021minimalist,abdolmaleki2021dime}. Our methods build on some of these: the exponential advantage-weighted actor-critic formulation from CRR~\cite{wang2020crr} and AWAC~\cite{nair2020awac}.

Building on the idea of speeding up RL by reusing stored data, several extensions to offline RL algorithms have been proposed which first derive a policy offline (from stored data) and then continue learning online by interacting with the environment. Algorithms in this class -- referred to as offline-to-online RL algorithms -- range from simple fine-tuning to more involved schemes that adapt some of their hyperparameters with the shifting data-distribution~\cite{nair2020awac,lee2021offlinetoonline}.
We note that the common setting for these approaches involves a large offline dataset given a priori, followed by fast adaption online. This differs from our setting where any offline and online data must come out of the shared data budget; we do not assume a large offline dataset.

Finally, a large part of our experiments builds on kickstarting algorithms, which are online RL algorithms aimed to be sped up by providing additional supervisory signals from suboptimal teachers.
This setting was originally made prominent for what is referred to as interactive imitation learning from an expert, or DAgger~\cite{ross2011dagger}, %
and was later adapted to the setting of kickstarting RL by using a \emph{suboptimal} teacher or prior policies to provide a supervision signal~\cite{schmitt2018kickstarting,parisotto2015actormimic}, including some works that considered using data collected from the teacher~\cite{cheng2018fast,jeong2020req}. This  setting can also be viewed from the perspective of RL as probabilistic inference (see e.g. \cite{levine2018reinforcement} for an overview) in which the RL objective is augmented to maximize reward while staying close to a prior policy. This prior can be instantiated with a suboptimal teacher policy, which is often used for transfer learning~\cite{tirumala2020behavior,abdolmaleki2021dime}.

Xie et al.~\cite{xie2021policyfinetuning} have recently contributed a theoretical analysis of the problem that they define as policy finetuning, where a policy for an MDP must be learned as efficiently as possible with access to a teacher policy. The analysis finds that depending on the distance from the teacher's state visitation distribution to the optimal policy's, the best policy finetuning algorithm is between ``offline reduction," i.e. offline RL from data collected by the teacher, and online RL. While we do find that offline reduction provides a strong baseline, for our realistic setting of learning the best policy with a limited data budget we find that mixing offline and online learning can perform better. 

\section{Preliminaries}

All algorithms we consider operate in the standard Markov Decision Process (MDP) framework, with states $s \in \mathcal{S}$, actions $a \in \mathcal{A}$, policy distribution $\pi(a | s)$, rewards $r(s, a)$, and discount factor $\gamma \in [0, 1)$. Without loss of generality, the rewards are sparse and the states are vision-based observations (two camera images and proprioception) in our experiments.
The goal of an RL algorithm is to find a policy $\pi$ that maximizes the discounted return $\EXP_{p_\pi}[\sum_{t=0}^\infty \gamma^t r(s_t, a_t)]$ under the trajectory distribution $p_\pi$ induced by the policy $\pi$.
We use $Q^\pi(s, a) = \EXP_{p_\pi} [\sum_{t=0}^\infty \gamma^t r(s_t, a_t) | s_0 = s, a_0 = a]$ to denote the Q-function
and $A^\pi(s, a) = Q^\pi(s, a) - \mathbb{E}_{a^\prime \sim \pi} [Q(s, a^\prime)]$ for the advantage.

\subsection{Online off-policy RL: Maximum a Posteriori Optimization}
\label{sec:mpo}
The first class of algorithms we build upon are RL algorithms trained by interacting online with the environment. We focus on off-policy actor-critic methods for  this setting as these are more data-efficient than on-policy counterparts (see e.g. \cite{haarnoja2018sac,abdolmaleki2018mpo}). In particular, we build upon the MPO algorithm~\cite{abdolmaleki2018mpo}, a state-of-the-art off-policy algorithm that, as we show below, can be adapted for the kickstarting setting and can also be related to offline RL algorithms~\cite{abdolmaleki2021dime}.

Learning in MPO is performed by alternating between policy evaluation and policy improvement steps, with concurrent data-collection to fill a replay buffer. The policy evaluation step at iteration $k$ aims to find an approximation $Q_{\phi_k}(\s, a) \approx Q^{\pi_k}(s, a)$ via temporal-difference learning (we use a distributional critic~\cite{bellemare2017distributional}). The policy-improvement step in iteration $k$ then aims to derive an improved policy by finding $\theta_{k+1} = \arg \max_\theta \text{MPO}(\pi_\theta, \pi_{k})$, with
\begin{equation}
    \text{MPO}(\pit, \pi_k) = \EXP_{\substack{s \sim \B \\ a \sim \pi_k}} \left[ \exp \left(\tfrac{1}{\eta}Q_{\phi_k}\!(\s, a) - Z\right) \log \pit(a | \s) \right],
    \label{eq:mpo}
\end{equation}
where $\B$ is the replay buffer, the expectation over actions can be approximated via $n$ samples from the policy $\pi_k$ and $Z = \log \EXP_{\pi_k}[\exp(\nicefrac{Q_{\phi_k}(\s, a)}{\eta})]$. This objective can be seen as minimizing the KL divergence between $\pit$ and an improved policy $\pi_\text{imp}(a  | \s) \propto \pi_k(a | \s) \exp(\nicefrac{Q_{\phi_k}(\s, a)}{\eta})$, where this improved policy is the solution of maximizing the KL-regularized objective: $\mathcal{J}(\pi_\text{imp};\pi_k) = \EXP_{\s,a}[Q_{\phi_k}(\s, a) - \eta \KLcompact{\pi_\text{imp}}{\pi_k}]$. In practice, rather than treating $\eta$ as a hyper-parameter, MPO optimizes it by considering a constrained optimization, choosing $\eta$ such that the KL is bounded by a user-defined $\epsilon$.

\subsection{Offline RL: Critic-regularized Regression}
\label{sec:crr}
Offline RL aims to learn a policy purely from a fixed dataset $\Dataset$ of transitions without further interaction. A major problem of applying standard off-policy RL algorithms (e.g. MPO) in this setting is that policy optimization (e.g. \autoref{eq:mpo}) can exploit the Q-function, due to bootstrapping in wrong estimates that cannot be corrected by observing true outcomes of the derived policy~\cite{fujimoto2019bcq,kumar2019bear}. In order to prevent this, some algorithms consider a KL-regularized optimization problem, regularizing towards the behavior policy $\pib$ that generated $\Dataset$. 
We consider CRR~\cite{wang2020crr} with exponential weights (equivalent to AWAC~\cite{nair2020awac} for offline RL), which maximizes the exponential advantage-weighted likelihood of actions in the dataset,
\begin{equation}
    \text{CRR}(\pit) \!=\! \EXP_{(\s, a) \sim \Dataset} \! \left[ \exp \! \left(\tfrac{1}{\eta}A_{\phi_k}\!(\s, a)\right)  \log \pit(a | \s) \right],\!
    \label{eq:crr}
\end{equation}
where $A_{\phi_k}(s, a) = Q_{\phi_k}(s, a) - \EXP_\pi[Q_{\phi_k}(s, a)]$ and where $\eta$ is now  a hyperparameter. This objective implicitly constrains the policy $\pit$ to only consider actions in the dataset and, similarly to MPO, can be seen as minimizing the KL divergence between $\pit$ and an improved version of the behavior policy $\pib$ given as $\pi_\text{imp}(a | \s)  \propto \pib(a | \s) \exp(\nicefrac{A_{\phi_k}(\s, a)}{\eta})$.

\section{Learning from suboptimal teachers by bridging kickstarting and offline RL}
\label{sec:method}
To enable data-efficient policy learning, we leverage known suboptimal solutions to the tasks we consider. We assume access to such a policy $\pie(a | \s)$ that we can use for data collection or to provide a supervision signal for our training procedure. We here consider $\pie$ to be the output of another learning algorithm trained, for example, via imitation of a teacher or via RL on a related problem. 

The main question we are concerned with then becomes: given a fixed data budget of $N$ trajectories (obtained by interacting with the environment), how should we best use $\pie$ to accelerate RL? Two obvious solutions are: i) we could use $\pie$ to collect $N$ trajectories and subsequently perform \textit{offline RL} to learn a policy which improves over $\pie$, or ii) we could perform standard online RL with a fixed budget of $N$ episodes and, optionally, query $\pie$ for additional supervision in each visited state, i.e. \textit{kickstarting}.

We present a partial answer to the questions of how to combine the best of these two strategies and how performance depends on the data budget $N$. 
To this end, we first derive variants of online and offline RL algorithms that can optionally use additional supervision from $\pie$. We achieve this for both of the algorithms we are building upon, MPO and CRR, by first realizing that they can be written in a unified general form as minimizing the KL divergence between the parametric policy $\pit$ and an improved policy $\pi_\text{imp}(a | \s) \propto \pip(a | \s) \exp(\nicefrac{Q_{\phi_k}(\s,a)}{\eta})$, where this improved policy is the solution of the KL-regularized RL problem with a general prior $\pip$: $\mathcal{J}(\pi_\text{imp};\pip) = \EXP_{\s,a}[Q_{\phi_k}(\s, a) - \eta \KLcompact{\pi_\text{imp}}{\pip}]$. The corresponding objective is the general exponential weighted problem,
\begin{equation}
    \hat{\mathcal{J}}(\pit; \pip) = \EXP_{\substack{s \sim \D \\ a \sim \pip}} \left[ \exp \left(\tfrac{1}{\eta} Q_{\phi_k}(\s, a) - Z\right) \log \pit(a | \s) \right]\!,
    \label{eq:eqw}
\end{equation}
from which we can recover MPO (modulo additional regularization used in MPO's update step) by setting $\D = \B$ and $\pip = \pi_k$, $Z = \log \EXP_{\pi_k}[\exp(\nicefrac{Q_{\phi_k}(\s, a)}{\eta})]$, and CRR by setting $\D = \Dataset$ and $\pip = \pib$, $Z = \mathbb{E}_{\pib} [\nicefrac{Q_{\phi_k}(\s, a)}{\eta}]$, restricting the sample estimate to actions from the dataset $\Dataset$. This formulation has also been studied in the literature when $\pip$ is not a previous policy or the behavior policy but a fixed (learned) prior; see e.g. \cite{tirumala2020behavior,galashov2020iwp} where pre-trained general priors were used to quickly adapt to new scenarios. We note that our contribution here is not a new training paradigm but a proposal for specific instantiations of $\pip$ (a mixture similar to DAgger/AggreVaTeD \cite{ross2011dagger,cheng2018fast}) to enable data-efficient policy finetuning, together with an extensive empirical evaluation.

If we want to endow MPO with additional supervision from the teacher $\pie$ we now can simply adjust $\pip$ to 
\begin{equation}
\textbf{R-MPO: } \pip(a | \s) = (1 - \beta) \pi_k(a | \s) + \beta \pie(a | \s),
\label{eq:prior_rmpo}
\end{equation}
in \autoref{eq:eqw}. We refer to this variant as MPO with relabeled actions, or R-MPO.
Similarly, we can endow CRR with additional supervision from the teacher by setting $\pip$ to
\begin{equation}
\textbf{R-CRR: } \pip(a | \s) = (1 - \beta) \pib(a | \s) + \beta \pie(a | \s),
\label{eq:prior_rcrr}
\end{equation}
i.e. a mixture of the behavior policy and the teacher.
We refer to this variant as CRR with relabeled actions, or R-CRR.
The behavior policy $\pib$ corresponds to the action distribution used for collecting the data $\D$.
In our setting, this is the teacher policy when the data is the offline dataset $\Dataset$, whereas it is the online policy when the data is the replay buffer $\B$, where the online policy is from the iteration when the data was added to the replay.
In our experiments, we allow for a mix of reusing fixed data collected by $\pie$ and the standard kickstarting setting, using the combined data $\D = \Dataset \cup \B$ for training, which we show to be critical to learning faster and achieving higher asymptotic performance.

\section{Experimental Setup}
\label{sec:experiments_setup}
Our experimental study compares a number of methods in the policy finetuning setting, with different teachers and data budgets, on a challenging robotic manipulation environment.
We first describe the components of this setup and then present results in the next section.

\subsection{Environment and Tasks}
As our primary interest is  efficiently learning policies in vision-based and dynamic robotic tasks, we chose
the RGB-stacking environment~\cite{lee2021beyond} as a case study due to its multi-object complexity, its requiring different behaviors for different shapes, and its matching environment versions in simulation and the real world.
The environment consists of a robotic arm, and three objects in a basket in front of the robot. 
The objects are colored red, green, and blue to signal to a vision-based agent which one should be stacked (red) on top of which (blue), and which one is just a distractor (green). These objects can be any of the $152$ RGB-objects, which were designed to vary the grasping and stacking affordances for a parallel gripper. The benchmark designates $5$ object triplets for evaluation, out of $3$ million possible triplets. \autoref{fig:triplet_challenges} depicts the 5 test triplets and demonstrates the unique challenges they each pose for stacking. 
We evaluate all methods we consider in this work on learning stacking using each of the test triplets individually, i.e. on $5$ distinct variations of the stacking task.%
\footnote{This is in contrast to Lee et al.~\cite{lee2021beyond} where skill mastery was evaluated on these triplets when training on all $5$ test triplets simultaneously.}

\subsection{Teacher Policies}
The ability of our methods to learn from a teacher will depend on how good the teacher is. For all $5$ stacking task variations we consider, we use $2$ different vision-based teacher policies, trained under each of the two settings outlined in the RGB-Stacking Benchmark~\cite{lee2021beyond}. These are: ``Skill Mastery'', under which a single teacher policy is trained to \textit{master} stacking on all $5$ test triplets jointly; and ``Skill Generalization'', under which a teacher learns by stacking objects from a designated training set, holding out the ones that comprise the test triplets. The generalization task is significantly more difficult and this teacher underperforms on the $5$ test triplets, both quantitatively, evident in \autoref{tab:teacher_performance}, and qualitatively, i.e. it demonstrates a lack of behavior variety. Both teacher policies were trained in simulation with domain randomization, which allows us to use the same teachers for simulation and real-robot experiments. 

\begin{table}
    \centering
    \begin{tabular}{@{\hspace{0.75\tabcolsep}}l@{\hspace{0.75\tabcolsep}}c@{\hspace{1\tabcolsep}}c@{\hspace{1\tabcolsep}}c@{\hspace{1\tabcolsep}}c@{\hspace{1\tabcolsep}}c@{\hspace{0.5\tabcolsep}}}
        \toprule
        Teacher                           & Triplet 1 & Triplet 2 & Triplet 3 & Triplet 4 & Triplet 5 \\
        \midrule
        Mastery (sim)                     &    78.9\% &    50.8\% &    82.8\% &    77.4\% &    86.7\% \\
        Generalization (sim)              &    36.5\% &    41.1\% &    44.7\% &    81.5\% &    90.4\% \\
        \hdashline \noalign{\vskip 0.5mm}
        Generalization (real)             &    39.9\% &    36.6\% &    39.5\% &    88.4\% &    83.4\% \\
        \bottomrule
    \end{tabular}
    \caption{Success rates of the mastery and generalization teachers, evaluated in simulation and on the real robots.
    The mastery teacher is not used for real-robot experiments. %
    }
    \label{tab:teacher_performance}
\end{table}
\begin{table}
    \centering
    \begin{tabular}{l c c@{\hspace{0.75\tabcolsep}}c}
        \toprule
        Methods      & Data $\D$                               & Prior policy $\pip$ \\
        \midrule
        BC           & $\Dataset$                              & $\pib$ \\

        CRR          & $\Dataset$                              & $\pib$ \\
        CRR-mixed    & $\Dataset \cup \B$                      & $\pib$ \\
        AWAC         & $\Dataset \rightarrow \Dataset \cup \B$ & $\pib$ \\
        DAgger       & $\B$                                    & $\pie$ \\
        DAgger-mixed & $\Dataset \cup \B$                      & $\pie$ \\
        MPO          & $\B$                                    & $\pi_k$ \\
        R-MPO        & $\Dataset \cup \B$                      & $(1-\beta) \pi_k + \beta \pie$ \eqref{eq:prior_rmpo} \\
        R-CRR        & $\Dataset \cup \B$                      & $(1-\beta) \pi_b + \beta \pie$ \eqref{eq:prior_rcrr} \\
        \bottomrule
    \end{tabular}
    \caption{Overview of the methods used in our experimental analysis.
    The objectives of all these methods can be unified under \autoref{eq:eqw}, with the differences being on the weighting of the log-likelihood, the state-action distribution of the data $\D$, and the action distribution of the prior policy $\pip$.
    All the methods use the exponential weighting, except for BC, DAgger, and DAgger-mixed.
    The first 4 methods do not use the behavior policy $\pib$ in queryable form; they only consider the actions in the data $\D$.
    }
    \label{tab:method_distributions}
\end{table}

\begin{figure*}
    \centering
    \begin{subfigure}[b]{\textwidth}
        \centering
        \includegraphics[width=\textwidth]{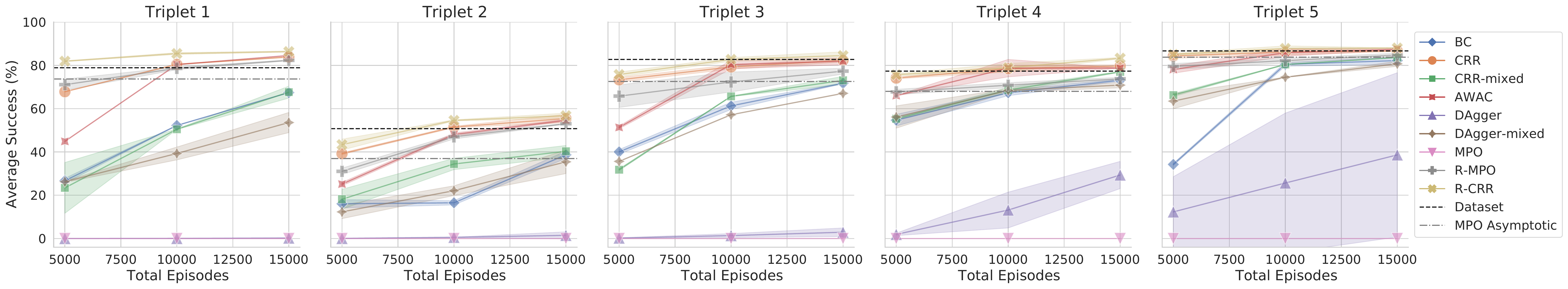} 
        \caption{Mastery teacher}
    \end{subfigure} \\[2mm]
    \begin{subfigure}[b]{\textwidth}
        \centering
        \includegraphics[width=\textwidth]{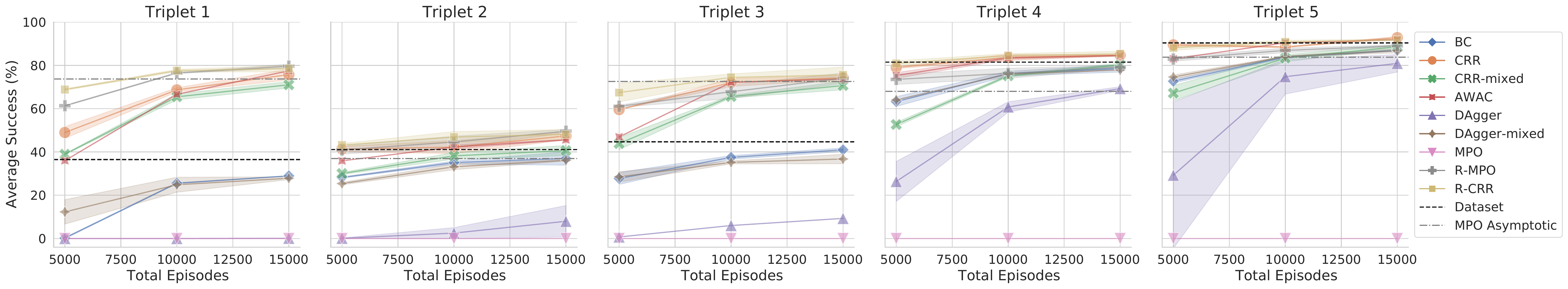} 
        \caption{Generalization teacher}
    \end{subfigure}
    \caption{Comparison of methods in simulation across data budgets, for each teacher and object triplet. The horizontal dashed lines indicate the performance of the teacher and the dash-dotted lines indicate the performance of MPO at \num{800000} episodes.
    MPO is not successful at stacking in under \num{15000} episodes, despite being independently tuned for the lower-data regime.
    The methods that use the combined sources of data, $\D = \Dataset \cup \B$, report the average success for the best proportion of offline episodes. 
    See \autoref{fig:sim_data_ratio_plot_generalist}  %
    for a comparison among different proportions of offline episodes.
    Our R-CRR method generally achieves the best performance while offline CRR is surprisingly competitive especially at the largest data budgets.
    }
    \label{fig:sim_episodes_plot}
\end{figure*}

\subsection{Methods}
The methods we chose for our comparisons all leverage an existing teacher policy that is suboptimal for the target task.
We consider methods that address the policy finetuning problem, from offline RL to kickstarting, standard online RL, and combinations of these.
The training objective of all the methods can be unified under \autoref{eq:eqw}, with the differences being on the weighting of the log-likelihood, the state-action distribution of the data $\D$, and the state-conditioned action distribution of the prior policy $\pip$.
\autoref{tab:method_distributions} summarizes the distributions used by each of the methods that we consider.

For the offline algorithms, the teacher policy is used to collect an offline dataset $\Dataset$ but it is not used afterwards during training.
Behavioral Cloning (\textbf{BC}) maximizes the likelihood of actions in the dataset: $\EXP_{(s, a) \sim \Dataset} [ \log \pit(a | s) ]$.
\textbf{CRR}~\cite{wang2020crr} instead weights the log-likelihood by the exponential advantage (\autoref{eq:crr}).
\textbf{CRR-mixed} is a simple variation of CRR, introduced here as a baseline, which incorporates online data by using the mixed source of data $\D = \Dataset \cup \B$, instead of purely training on the offline dataset.
\textbf{AWAC}~\cite{nair2020awac} is an offline-to-online RL algorithm that first trains the policy offline from the fixed dataset, and then transitions to online training.
Effectively, it is like CRR, except that the algorithm transitions from using only $\Dataset$ to using mostly $\B$ throughout the duration of training. %
These four methods implicitly use the behavior policy $\pib$ for the prior when the data is the dataset, but the expectation over the prior is approximated via a single action sample for each state -- the state-action pair in the dataset.

For kickstarting, the teacher policy is used to supervise the actions for the states visited by the student.
\textbf{DAgger}~\cite{ross2011dagger} optimizes the same loss as BC except that the actions are queried from the teacher on states from trajectories obtained by executing the student policy: $\EXP_{s \sim \D, a \sim \pie} [ \log \pit(a | s) ]$ with $\D = \B$.
\textbf{DAgger-mixed}, also introduced here as a baseline, instead trains on the mixed source of data from the teacher and the student, $\D = \Dataset \cup \B$, which we found to be more sample-efficient than standard DAgger.
At the far end of the spectrum, \textbf{MPO}~\cite{abdolmaleki2018mpo} is a fully online RL algorithm that does not use teacher supervision, so we instead trained it with shaped rewards (only for MPO).

Lastly, the methods we introduce in this work in \autoref{sec:method}, \textbf{R-MPO} and \textbf{R-CRR}, use the exponential advantage-weighted log-likelihood objective from \autoref{eq:eqw}, where the actions are sampled from a prior policy. The prior is a mixture of the target and teacher policies in R-MPO (\autoref{eq:prior_rmpo}), or the behavior and teacher policies in R-CRR (\autoref{eq:prior_rcrr}).

\begin{figure*}
    \centering
    \includegraphics[width=\textwidth]{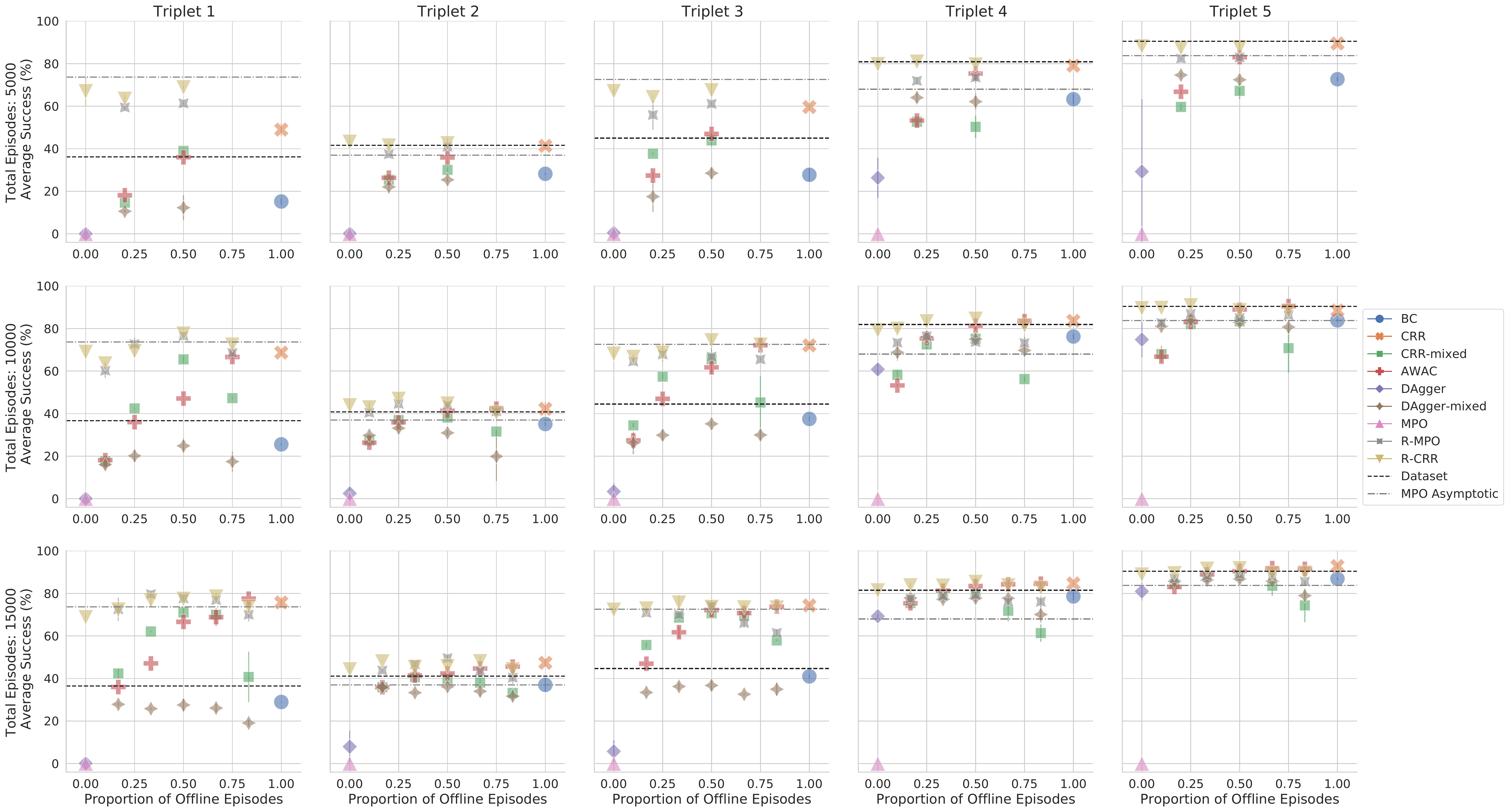} 
    \caption{
    Comparison of methods in simulation across proportions of offline episodes from the total budget, for the generalization teacher, for each data budget and object triplet. The horizontal dashed lines indicate the performance of the teacher and the dash-dotted lines indicate the performance of MPO at \num{800000} episodes.
    The data proportions were based on using offline datasets of sizes \num{1000}, \num{2500}, \num{5000}, \num{7500}, \num{10000}, \num{12500}, and \num{15000}. Different methods have different sensitivities to this parameter. R-CRR generally works best when half of the data budget are offline episodes from the teacher.
    CRR does not benefit from mixing in online episodes (CRR-mixed) whereas DAgger significantly benefits from mixing in offline episodes (DAgger-mixed).
    }
    \label{fig:sim_data_ratio_plot_generalist}
\end{figure*}

\subsection{Data budgets \& proportions of offline data}
We compare these methods over several data budgets, i.e.
the \textit{total} number of episodes -- both offline and online -- collected in the target task for the purposes of training the resulting policy $\pit$. For the methods which use a combination of offline episodes in the dataset $\Dataset$ and online episodes in the replay buffer $\B$, we can vary the proportion of offline episodes within the total budget. We obtain different offline proportions by generating fixed datasets $\Dataset$ of different sizes from teacher rollouts and using the remaining budget on episodes from online training.

\section{Experimental Results}
\label{sec:experiments}
We compare a number of methods in the policy finetuning setting, with different teachers and data budgets, on 5 versions of a challenging robotic manipulation environment, both in simulation and the real world. 
We structure our experiments to answer the following research questions:
\begin{enumerate}
    \item How do different methods compare at each data budget in the policy finetuning problem setting?
    \item Is there a benefit of using offline episodes from the teacher during online training and, if so, how much?
    \item How much action supervision from the teacher suffices in the prior mixture?
\end{enumerate}

\subsection{Results in Simulation}
\label{sec:experiments_simulation}
In simulation, we compare all methods across the 5 vision-based stacking task variations, using both the mastery and generalization teachers, and 3 data budgets: \num{5000}, \num{10000}, and \num{15000} episodes. These budgets were selected to show the differences between methods, as CRR, AWAC, R-MPO, and R-CRR typically converged to similar performance by \num{15000} episodes. For the methods that use both offline and online data, we additionally vary the relative proportions of offline data from the teacher and online data from the student.
All reported numbers are an average from 2 runs (i.e. seeds) and evaluating each over \num{1000} episodes in each target task.

\subsubsection{Comparison of data budgets and proportion of offline episodes}
We summarize our results in \autoref{fig:sim_episodes_plot}, where we show a comparison of the performance of all  methods across the different data budgets on each of the 5 tasks and 2 teacher policies separately, while only reporting the best proportion of offline data for each combination of method and data budget.
CRR outperforms the teacher (shown by the horizontal dashed lines) in most cases, especially at the higher data budgets, without needing to collect any data online.
Our R-CRR method generally achieves the highest performance, with the difference relative to the other methods being larger at the lower data budgets and when the teacher is less optimal.
When the teacher performance is already relatively high (e.g. when using the mastery teacher, or when training on Triplets 4 or 5), the improvement over the teacher is smaller and most of the methods perform fairly well for the largest data budget of \num{15000} episodes.
R-MPO generally performs slightly worse than R-CRR and even offline CRR in most cases despite having access to online episodes.
Overall, the performance of the methods improves with more data, with the largest improvement observed when increasing the budget from \num{5000} to \num{10000} episodes. At the largest  budgets, offline RL with CRR is surprisingly competitive.

In \autoref{fig:sim_data_ratio_plot_generalist}, we look at performance  over different proportions for each budget.
We find that different methods have different sensitivities to this offline-online tradeoff, but that overall the best choice is to use half of the budget on offline teacher data, and half on online student data. %

\begin{figure*}
    \centering
    \begin{subfigure}[b]{0.5\textwidth}
        \centering
        \includegraphics[height=40mm]{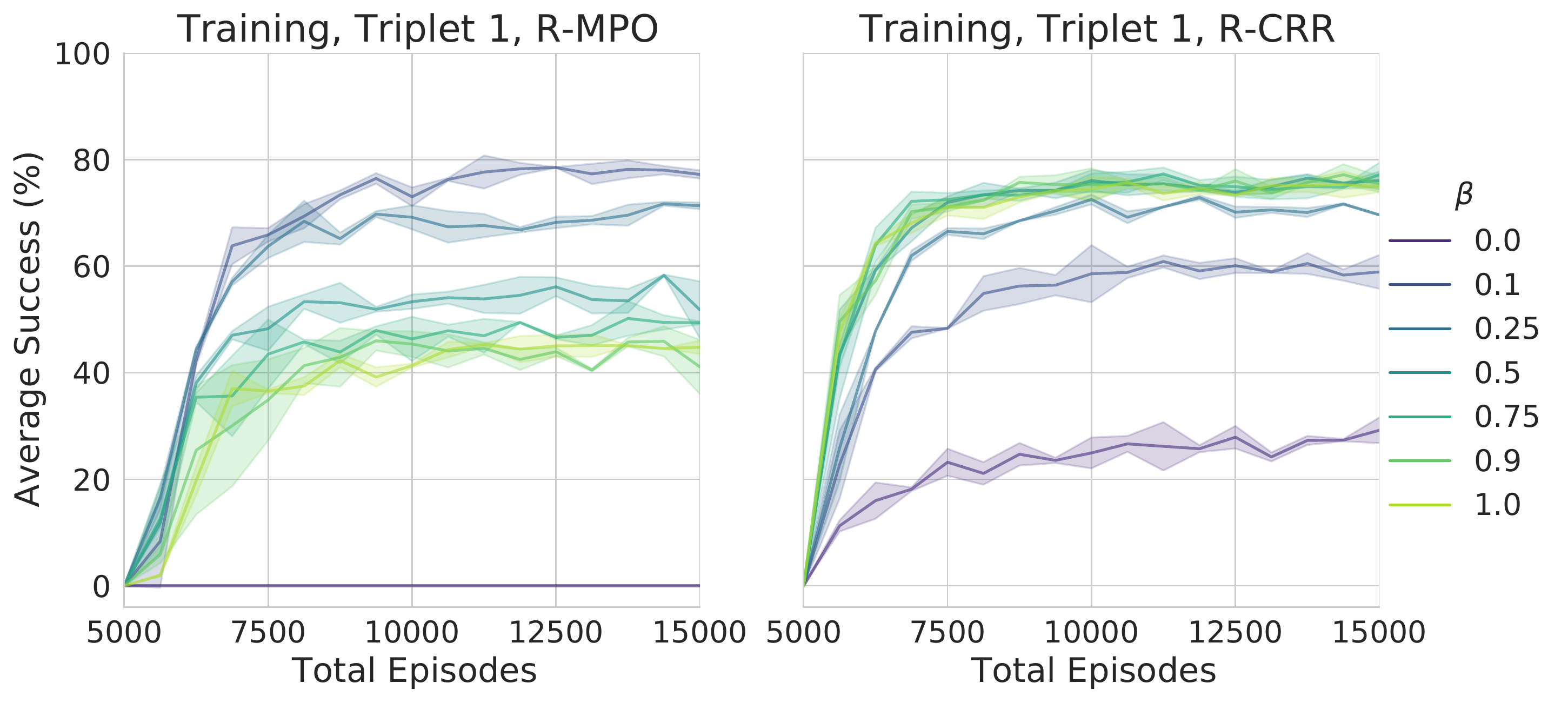} 
        \caption{Mixture weight $\beta$ for the teacher component of the prior policy}
        \label{fig:betaplot}
    \end{subfigure}\hfill%
    \begin{subfigure}[b]{0.2255\textwidth}
        \centering
        \includegraphics[height=40mm]{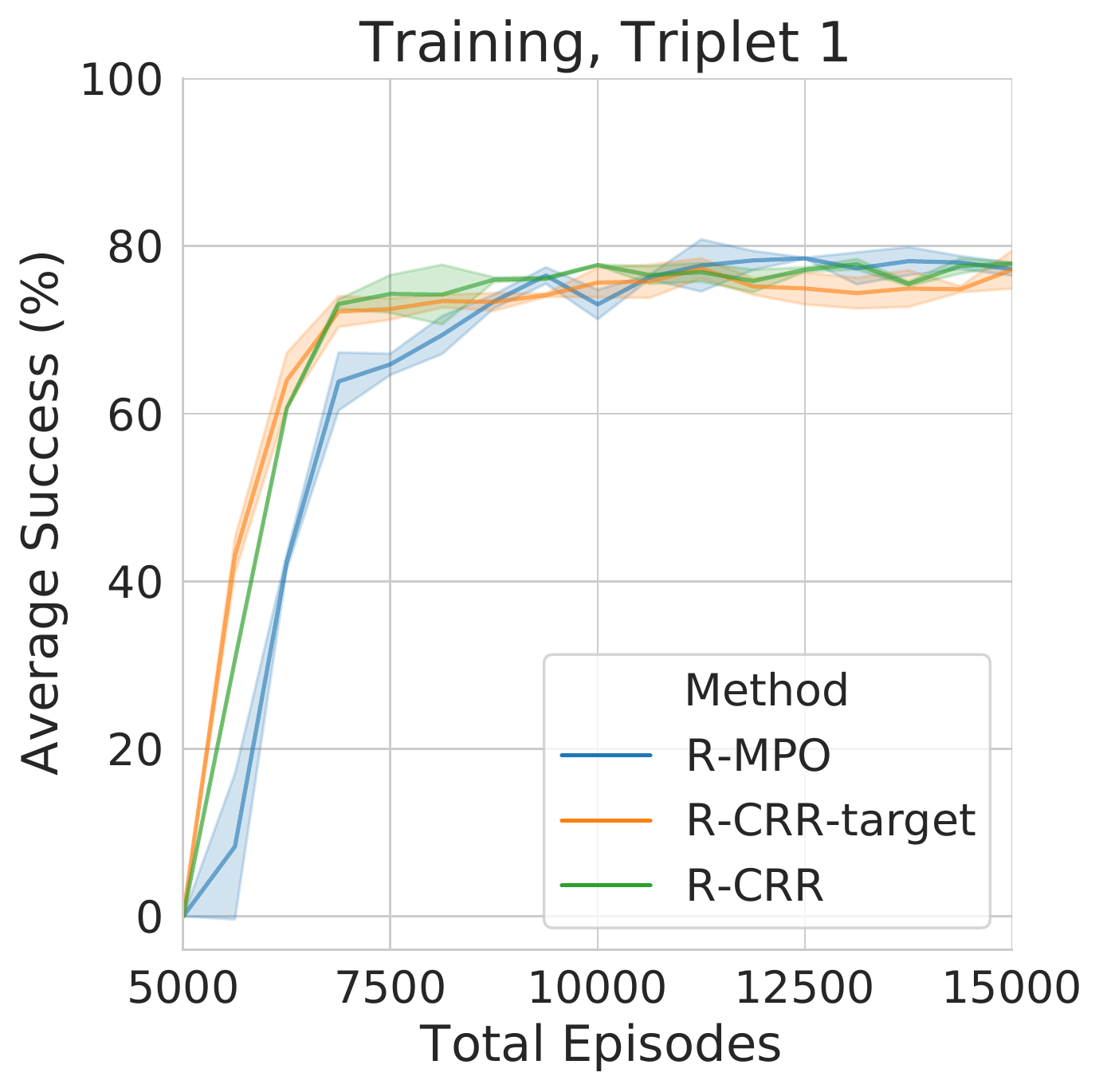} 
        \caption{Our method variants}
        \label{fig:kickstartingplot}
    \end{subfigure}\hfill%
    \begin{subfigure}[b]{0.27\textwidth}
        \centering
        \includegraphics[height=40mm]{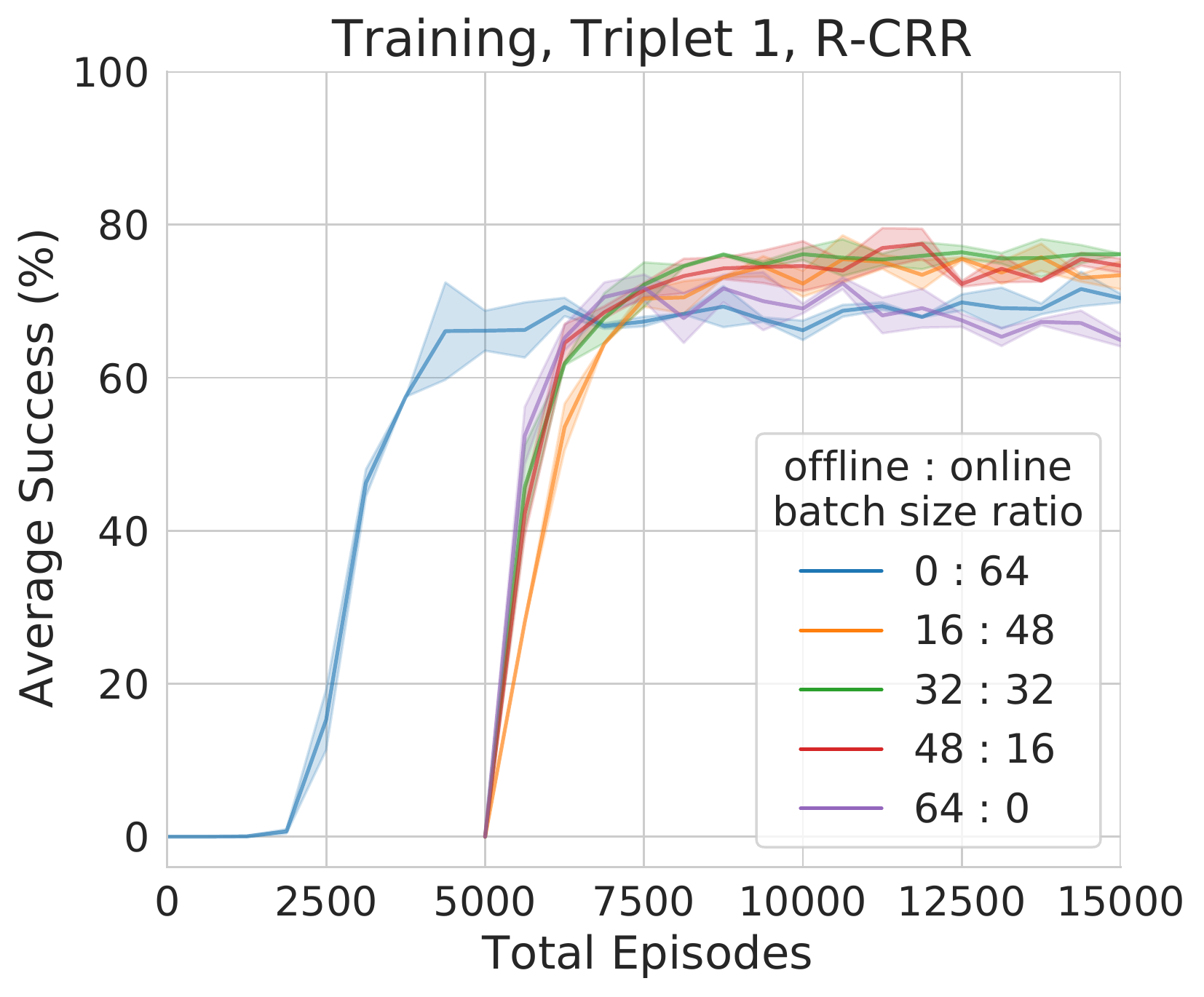} 
        \caption{Batch-size ratio}
        \label{fig:batchratioplot}
    \end{subfigure}
    \caption{Comparison, in simulation, of algorithmic choices and hyperparameters for our methods on Triplet 1 from the generalization teacher.
    These methods use the mixed source of data, $\D = \Dataset \cup \B$, where the dataset is fixed with \num{5000} offline episodes from the teacher, unless otherwise noted.
    \subref{fig:betaplot} R-MPO is better with a low non-zero weighting $\beta$ of the teacher within the prior mixture, which prevents constraining the policy too much, whereas R-CRR benefits from a higher weighting.
    \subref{fig:kickstartingplot} Our methods perform equally well asymptotically, though R-MPO learns slightly slower compared to R-CRR and its variant R-CRR-target.
    \subref{fig:batchratioplot} The batch-size ratio controls how many subsequences in each mini-batch are sampled from $\Dataset$ and $\B$. %
    R-CRR is not sensitive to this hyperparameter as long as the batch contains both sources of data.
    The ``\num{0} : \num{64}'' batch-size ratio is the only curve among these plots that do not use any offline episodes.
    }
    \label{fig:ablation}
\end{figure*}

\subsubsection{Hyperparameter analysis}
We show the effect of algorithmic choices and hyperparameters for our methods in \autoref{fig:ablation}.
We focus this study on the Triplet 1 task with the generalization teacher, as this task has a large potential for improvement from the teacher.
We fix the number of offline episodes to $5000$ and report learning curves, thus showing the performance for a continuum of total episodes.

As described in \autoref{sec:method}, our algorithms use a prior policy with a hyperparameter $\beta$, which controls how much to weigh the teacher's action distribution $\pie$ within the prior $\pip$. $\beta$ is independent of the proportion of offline teacher episodes, and rather affects how much $\pit$ can diverge from $\pie$.
As shown in \autoref{fig:betaplot}, R-MPO is best at $\beta=0.1$, while R-CRR benefits from higher values ---we used $\beta=0.75$ in all of our other experiments.
Notice that R-MPO with $\beta=0$ is similar to MPO, which struggles on long-horizon sparse-reward tasks, like most online RL algorithms would. Interestingly, a mixing weight as low as $\beta=0.1$ provides enough supervision to solve the task, while higher weights constrain the student too much, possibly due to the interplay with respect to the MPO KL constraint. %

Aside from the choice of the prior mixture in R-MPO (\autoref{eq:prior_rmpo}) and R-CRR (\autoref{eq:prior_rcrr}),
another difference is that the R-MPO method optimizes the \textit{KL-constrained} objective for a given constraint threshold $\epsilon$, whereas R-CRR optimizes the \textit{KL-regularized} objective treating $\eta$ as a hyperparameter.
We can consider an R-CRR variant, referred to as R-CRR-target, which also optimizes the KL-regularized objective but instead uses the prior policy as in R-MPO, i.e. a mixture of the target and teacher policies.
\autoref{fig:kickstartingplot} compares the learning curves of these three methods. 
Although these perform equally well asymptotically, R-MPO takes longer to improve.
We hypothesize that these KL-regularized methods with a fixed $\eta$ are less sensitive to the exact composition of the prior policy as long as the teacher proportion is high enough, which was the case when varying $\beta$ in \autoref{fig:betaplot}.

The methods using the mixed source of data, $\D = \Dataset \cup \B$, use mini-batches with half offline teacher data and half online data by default. \autoref{fig:batchratioplot} shows the effect of this ratio for R-CRR.
Note that this batch ratio is different and independent of the proportion of offline episodes in the data budget.
We found that as long as the batch contains both sources of data, the exact ratio in the batch does not matter for final performance.
We posit that training fully online (the ``0 : 64'' ratio) gets lower asymptotic performance due to overfitting to the replay buffer, which is small at the beginning of training as the replay is progressively being filled at a fixed rate as the agent performs training updates.
Despite this being the standard online RL paradigm, we argue that including a large set of data from the beginning of training, such as that of a fixed dataset with pre-collected episodes, can mitigate this issue.

\subsection{Results on the Real Robots}
Our real-robot experiments differ in a few ways from the simulation experiments due to resource constraints.
First, we only consider the generalization teacher as its lower performance allows for greater variation across methods. Second, the offline data is collected from the teacher deterministically (such as for evaluation) -- the mode action from the policy's output distribution is executed rather than a sampled action. However, the online algorithms use the sampled actions from $\pit$ so as to not limit the exploration of the student.

All reported numbers are an average of 2 training runs for the offline methods and only 1 training run for the online methods, and each evaluation is an average over \num{200} episodes per robot.
It takes just under a day to execute \num{1000} episodes on each robot, regardless whether for dataset collection, online training, or evaluation.

\begin{figure}
    \centering
    \includegraphics[height=40mm]{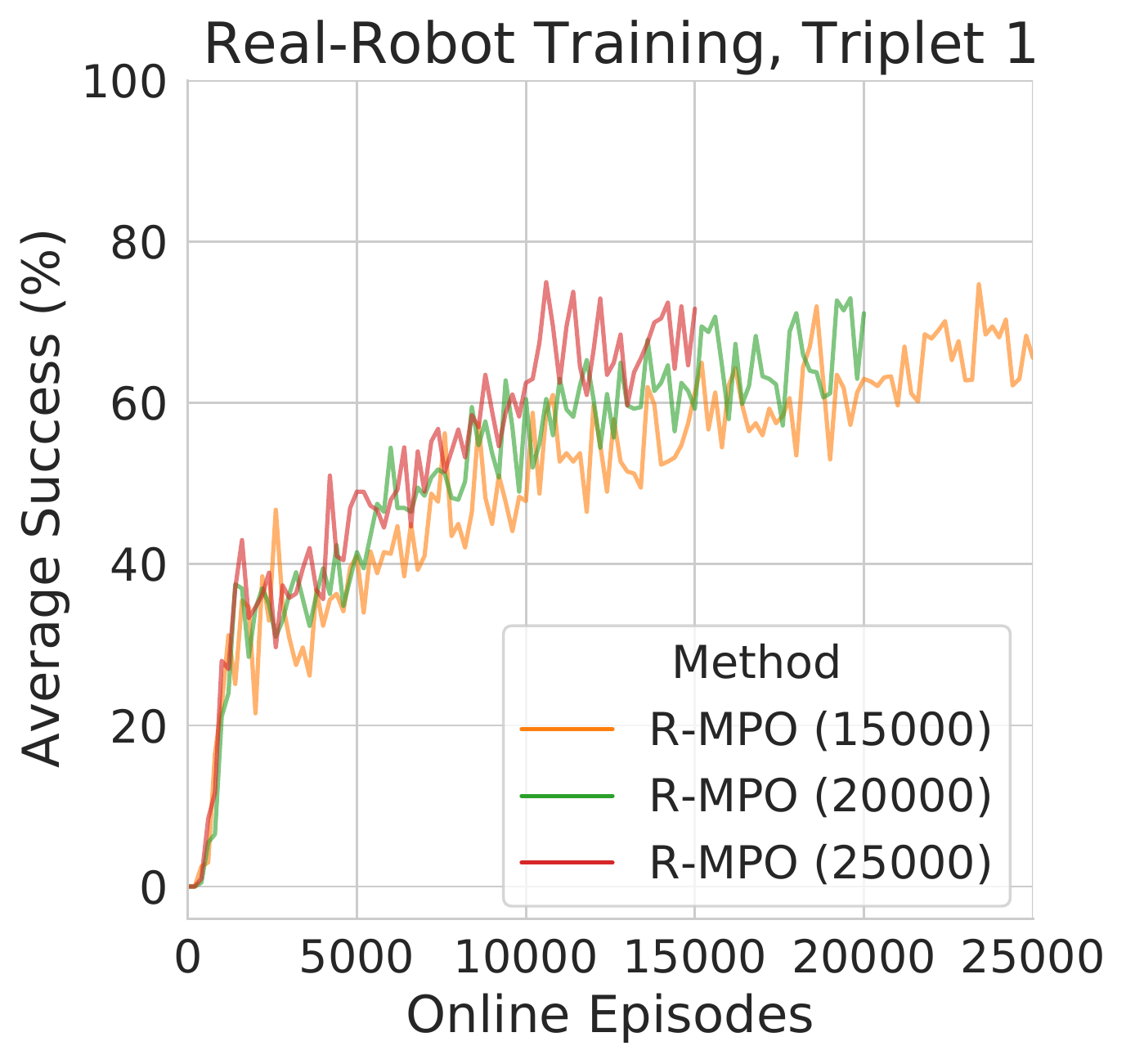}%
    \hfill%
    \includegraphics[height=40mm]{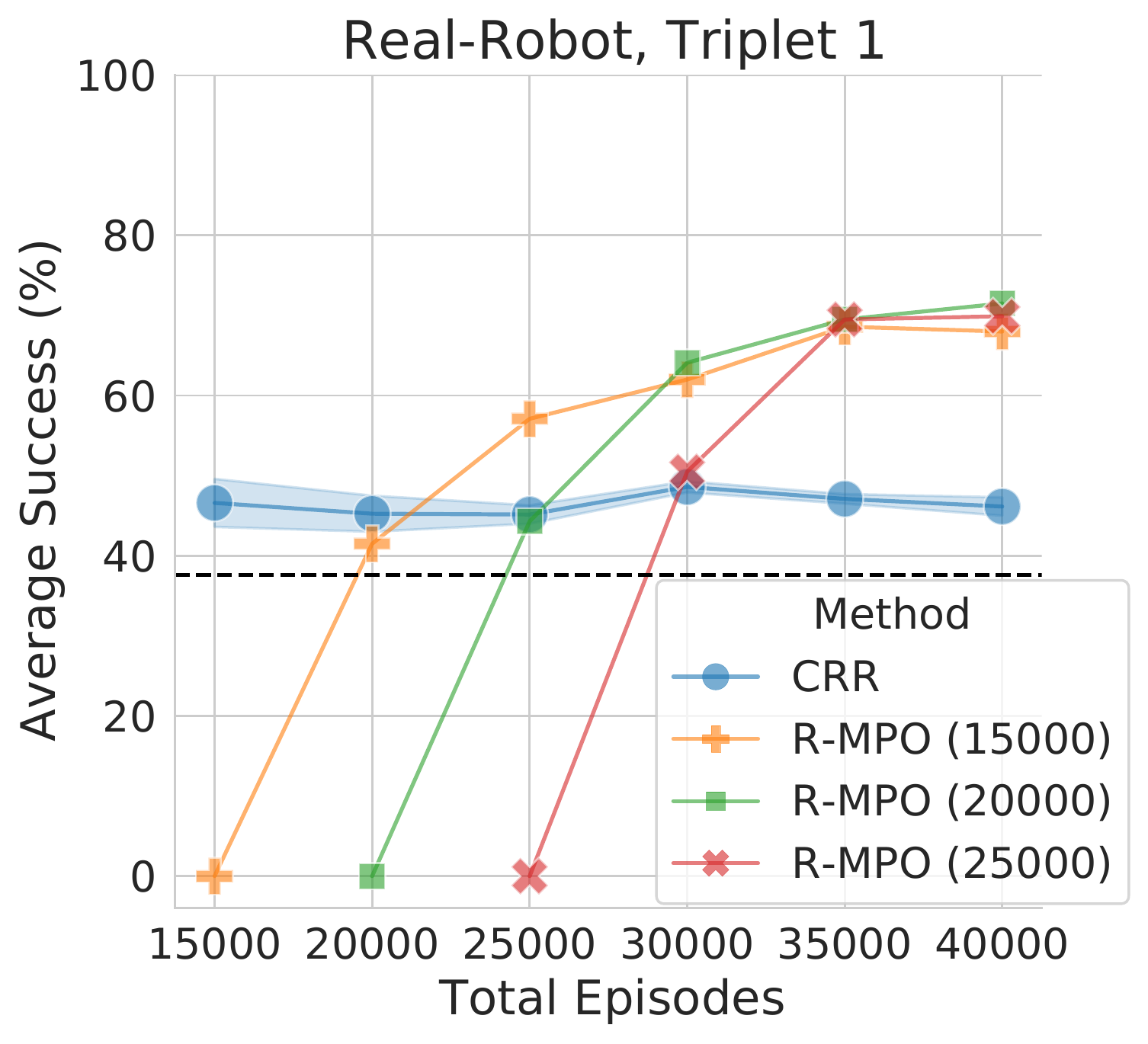}
    \caption{Real-robot comparison of CRR and R-MPO on Triplet 1, using the generalization teacher.
    The CRR runs were trained offline from datasets of various sizes.
    The R-MPO runs trained online on the robots for a total data budget of \num{40000} episodes, starting with \num{15000}, \num{20000}, or \num{25000} offline episodes.
    The left plot shows the learning curves for R-MPO, reporting the average success (over a window of the past 200 episodes) of the episodes executed by the stochastic online policy throughout training.
    The right plot shows the evaluation performance for CRR and R-MPO for a number of data budgets, reporting the average success of 200 episodes from the deterministic policy.
    The horizontal dashed line indicates the teacher performance, taken from the largest dataset used.
    }
    \label{fig:robot_learning_curve_rmpo_offlineepisodes_group1}
\end{figure}
\begin{figure*}
    \includegraphics[width=\textwidth]{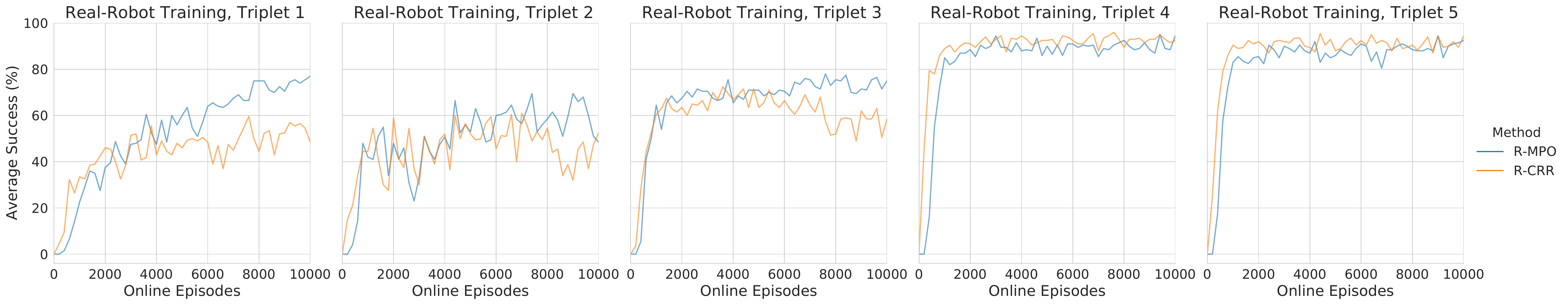} 
    \caption{Training curves of real-robot comparison of R-MPO and R-CRR for each object triplet using the generalization teacher. The methods were trained online on the robots for a total data budget of \num{35000} episodes, starting with \num{25000} offline episodes.
    We report the average success (over a window of 200 episodes) of the episodes executed by the stochastic policy throughout training.
    }
    \label{fig:robot_learning_curve_objectset_method}
\end{figure*}

\subsubsection{Comparison of data budgets}
Our first set of robot experiments compares offline RL and R-MPO on Triplet 1 for various data budgets and offline proportions.
Through preliminary experiments, we found that all methods perform poorly at data budgets below \num{15000} and that, in contrast to the simulation experiments, R-MPO outperformed R-CRR on Triplet 1. %
Because of these findings, we consider data budgets \num{15000} to \num{40000} in increments of \num{5000} episodes to compare R-MPO to offline CRR. 
We perform three separate training runs for R-MPO, each using a different number of offline episodes -- \num{15000}, \num{20000}, and \num{25000}.
We used 4 robots throughout this set of experiments -- for collecting offline episodes, online training, and for evaluation -- using equal numbers of episodes from each robot.  %

\autoref{fig:robot_learning_curve_rmpo_offlineepisodes_group1} shows learning curves for R-MPO and evaluation results at each data budget.
R-MPO significantly outperforms the teacher and its performance is equally good at the largest data budgets, starting to plateau at \num{35000} episodes, regardless of the proportion of offline episodes.
Although CRR improves upon the teacher, its performance is worse than R-MPO, without showing an upward trend with more data.
We hypothesize that CRR is negatively impacted from training on the datasets that, in our case, only contains episodes from the deterministic policy, lacking action diversity.

\subsubsection{Comparison of CRR, R-MPO, and R-CRR for each object triplet}
Our second set of experiments compares offline CRR and our methods -- R-MPO and R-CRR -- on all five object triplets. Due to resource constraints, we use a fixed total budget of \num{35000} episodes.
R-MPO and R-CRR used \num{25000} offline episodes and converged within \num{10000} online episodes (much less for some triplets). 
The object triplets were each assigned one robot, which remained constant throughout the entirety of the experiment.

\autoref{fig:robot_learning_curve_objectset_method} shows learning curves for R-MPO and R-CRR and \autoref{tab:robot_evaluation_objectset_method} shows the final performance at the \num{35000} data budget, for each of the 5 object triplets.
R-MPO is significantly better than the other two methods on Triplet 1, though there is no method that consistently outperforms the others on all the triplets. CRR performs surprisingly well on some of the triplets, despite only using offline data.
All the three methods outperform the teacher.

\begin{table}
    \centering
    \begin{tabular}{l c c c c c}
        \toprule
        Method      &  Triplet 1 &  Triplet 2 &  Triplet 3 &  Triplet 4 &  Triplet 5 \\
        \midrule
        CRR         &     48.3\% &     58.5\% &     61.0\% & \bf 94.5\% &     88.5\% \\
        R-MPO       & \bf 81.5\% &     54.5\% & \bf 73.0\% &     91.5\% &     89.0\% \\
        R-CRR       &     54.0\% & \bf 60.5\% & \bf 73.5\% & \bf 93.5\% & \bf 94.0\% \\
        \bottomrule
    \end{tabular}
    \caption{
    Evaluation success rates for offline CRR and our methods, R-MPO and R-CRR, on the real robots. We trained for each of the 5 triplets, using the generalization teacher and a data budget of \num{35000} total episodes. CRR offline used \num{35000} offline episodes. Both of our methods used \num{25000} offline episodes and \num{10000} online episodes.
    These evaluations used 200 episodes from the deterministic policy to compute every success rate.
    }
    \label{tab:robot_evaluation_objectset_method}
\end{table}

\section{Conclusion}
In this work we conducted a thorough experimental study of a number of methods for the policy finetuning problem setting on a set of stacking tasks, including offline, online, offline-to-online, and kickstarting RL algorithms.
We found that when a suboptimal teacher policy is available, using that teacher to collect a dataset and then performing offline RL (such as with CRR) is a simple and strong baseline that easily outperforms the teacher.
We also showed that using both offline episodes from the teacher and online episodes from the student policy, while also using the teacher to relabel actions for those states, leads to even better performance with our R-CRR algorithm (and our R-MPO algorithm in some cases), especially for the smaller data budgets.
As our results hold across 5 object triplets, 2 teacher policies, and in both simulation and real environments, we believe that our study will provide valuable lessons for other real-world robotic tasks.

\bibliographystyle{IEEEtran}
\bibliography{references}

\newpage
\appendices

\input{appendix}

\end{document}

%% file: appendix.tex
\section{Additional experiment details}
\label{app:experiments}
\begin{figure*}[t]
    \centering
    \includegraphics[width=\textwidth]{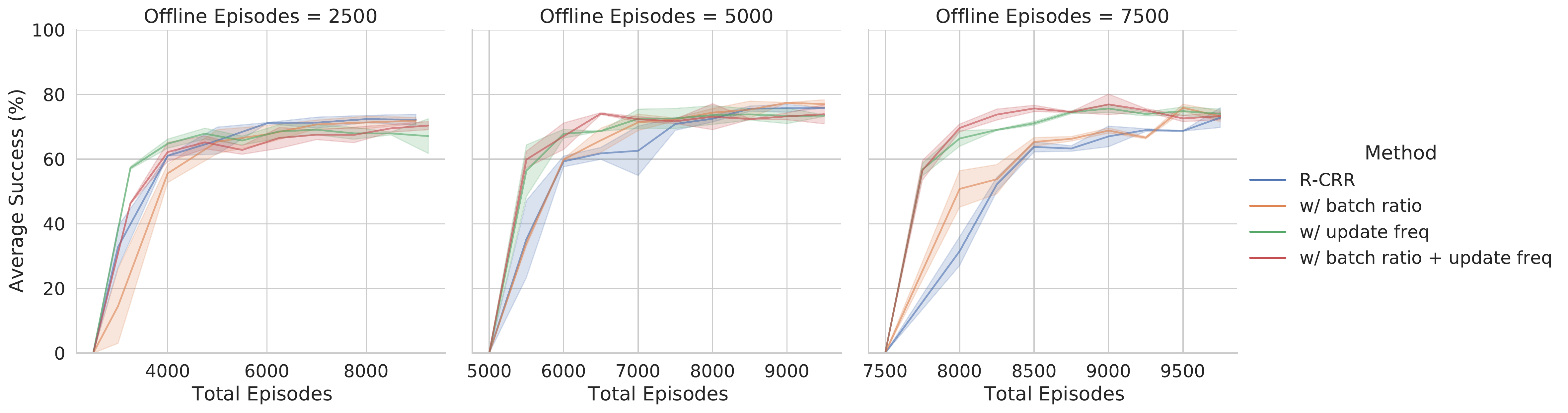}
    \caption{Ablation over batch ratio and gradient steps. In blue: our usual hyperparameters (batch that is half teacher data and half online student data, and 1 gradient step for every 5 timesteps). In orange: adjusting the rate of gradient steps per timestep (update frequency) based on the number of offline episodes such that the agent has trained for 1M gradient steps at the 10k total episodes. In green: changing the batch ratio of teacher to student data based on the total data ratio. In red: both modifications together. All are R-CRR on object group 1 with the generalization teacher and 10k data budget. These hyperparameters do not affect final performance.}
    \label{fig:spf_plot}
\end{figure*}
\begin{figure*}[t]
    \centering
    \includegraphics[width=\textwidth]{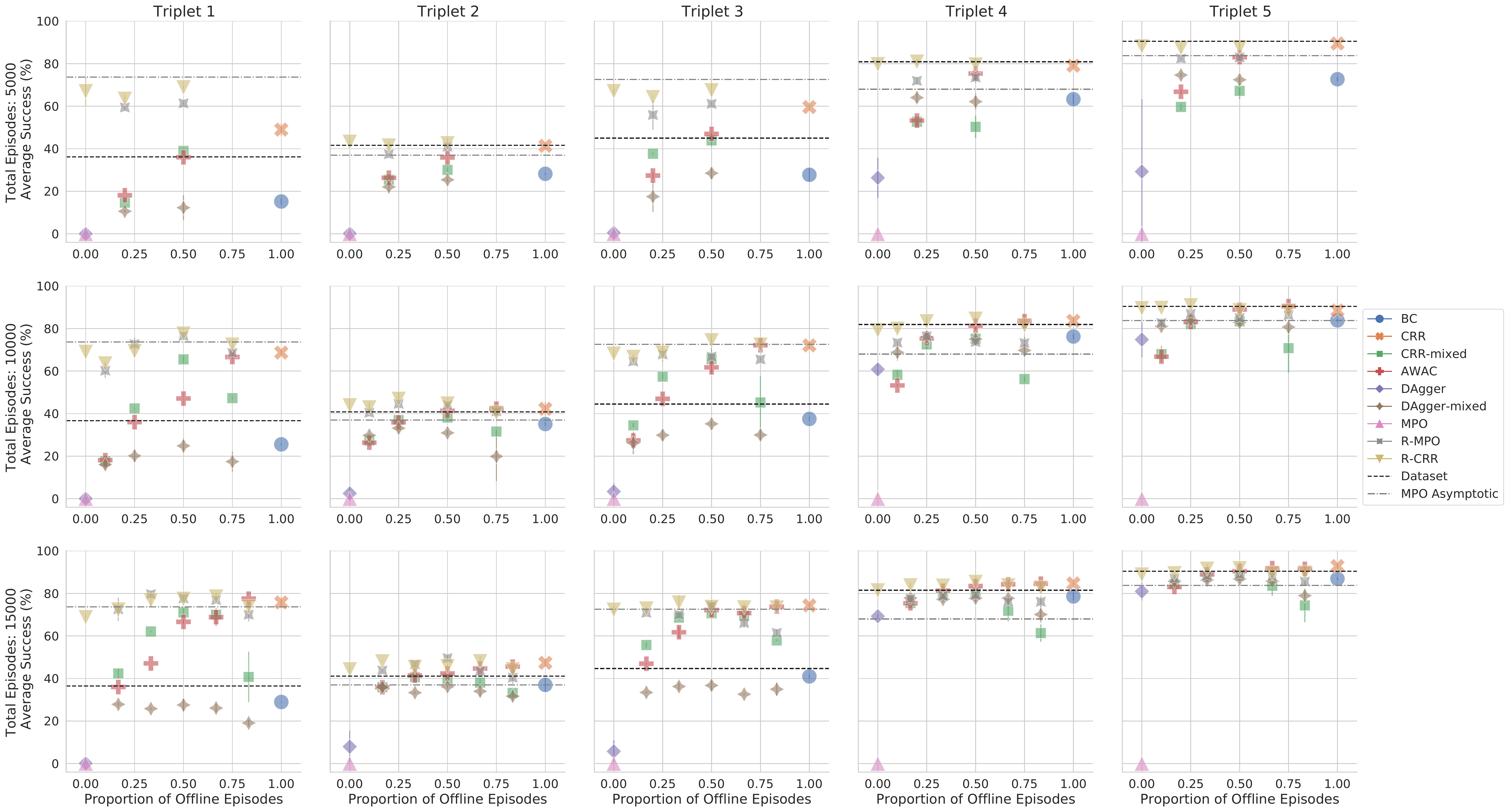} 
    \caption{
    Comparison of methods in simulation across proportions of offline episodes from the total budget, for the mastery teacher, for each data budget and object triplet. The horizontal dashed lines indicate the performance of the teacher and the dash-dotted lines indicate the performance of MPO at \num{800000} episodes.
    The data proportions were based on using offline datasets of sizes \num{1000}, \num{2500}, \num{5000}, \num{7500}, \num{10000}, \num{12500}, and \num{15000}. Different methods have different sensitivities to this parameter. Because the mastery teacher performs much better than the generalization teacher, high proportions of data from the teacher are generally quite effective.
    This figure is analogous to \autoref{fig:sim_data_ratio_plot_generalist} but for the mastery teacher instead of the generalization teacher.
    }
    \label{fig:sim_data_ratio_plot_mastery}
\end{figure*}

\paragraph{Data Budgets} In the simulation experiments, we compare the methods over 3 different data budgets: \num{5000} episodes, \num{10000} episodes, and \num{15000} episodes. Depending on the method, these episodes may be sampled offline from the teacher policy, online from the student, or from a mix of both.

For the offline and the mixed-data methods, which use teacher data, datasets are pre-collected from the teacher policies so that all experiments for each data budget use the same teacher data. Fully offline methods are all trained for 1M gradient steps while the methods using any online data are trained until the data budget is met. All methods use a batch size of 64.

For the mixed-data methods that use both online and offline data, we sample batches that are half from the offline dataset of $\pie$ episodes and half from the replay buffer being filled by the online student $\pit$. We keep fixed the rate of gradient steps per timestep -- 1 gradient step is taken for every 5 timesteps from the online actors. This means that the mixed-data methods that use more offline data end up training with fewer gradient steps.
We could instead keep fixed the total number of gradient steps by varying the rate of gradient steps per timestep. However, \autoref{fig:spf_plot} shows that matching the total number of gradient steps for the same data budget does not affect results compared to our main results with the fixed rate of gradient steps per timestep.

\paragraph{AWAC Settings}
Our AWAC agent is implemented on top of our CRR agent, with the main difference being that the training data first consists of data from the offline dataset $\Dataset$ and then gradually mixes in data from the online policy $\pib$ by increasing the proportion of data coming from the replay buffer $\B$ throughout the training. For all AWAC experiments, all the data in the batch is sampled from the offline dataset for the initial 500k gradient steps. Then we linearly decrease the offline-to-online data proportion from 100\% offline to 20\% offline and 80\% online throughout the next 600k gradient steps, while the replay buffer is being filled by episodes from student policy. The replay buffer starts to be filled at 500k gradient steps. In the meantime, the temperature is linearly decreased from 1 to 0.1 starting from 400k gradient steps and ending at 500k gradient steps. Those hyperparameters are chosen based on a comprehensive parameter sweep. Unlike the offline methods that trained for a fixed number of gradient steps, the AWAC methods are trained until the total data budget is met. The total gradient steps varies from approximately 700k to 1.6M depending on the data budgets.

\section{Additional Results}
\label{app:experiments_results}
In \autoref{fig:sim_data_ratio_plot_mastery} we show the full data ratio results in simulation using the mastery teacher. As compared to the generalization teacher results shown in \autoref{fig:sim_data_ratio_plot_generalist}, most methods (except for CRR-mixed) do not significantly drop off in performance with more than 50\% data from the teacher. R-CRR still usually performs best with a 1:1 data ratio and often outperforms even this high-performing mastery teacher.